\documentclass[letterpaper]{article} 
\usepackage{main2026}  
\usepackage{times}  
\usepackage{helvet}  
\usepackage{courier}  
\usepackage[hyphens]{url}  
\usepackage{graphicx} 
\usepackage{natbib}  
\usepackage{caption} 
\usepackage{algorithm}
\usepackage{algpseudocode} 
\usepackage{subcaption} 
\usepackage{amsmath}
\usepackage{amssymb}

\usepackage{newfloat}
\usepackage{listings}
\DeclareCaptionStyle{ruled}{labelfont=normalfont,labelsep=colon,strut=off} 
\floatstyle{ruled}
\newfloat{listing}{tb}{lst}{}
\floatname{listing}{Listing}
\usepackage{booktabs}       
\usepackage{amsfonts}       
\usepackage{nicefrac}       
\usepackage{microtype}      
\usepackage{xcolor}         
\usepackage{array}
\usepackage{textcomp}
\usepackage{verbatim}
\usepackage{extarrows}
\usepackage{multirow}
\usepackage{amssymb}
\usepackage{soul}
\usepackage{makecell}
\usepackage{pifont}
\usepackage{enumitem}
\title{Actor-Critic for Continuous Action Chunks: A Reinforcement Learning Framework for Long-Horizon Robotic Manipulation with Sparse Reward}
\author {
    Jiarui Yang\textsuperscript{\rm 1},
    Bin Zhu\textsuperscript{\rm 2},
    Jingjing Chen\textsuperscript{\rm 3}\thanks{Corresponding author.}, 
    Yu-Gang Jiang\textsuperscript{\rm 3}
}
\affiliations {
    \textsuperscript{\rm 1}College of Computer Science and Artificial Intelligence, Fudan University\\
    \textsuperscript{\rm 2}Singapore Management University\\
    \textsuperscript{\rm 3}Institute of Trustworthy Embodied AI, Fudan University\\
    jryang24@m.fudan.edu.cn, binzhu@smu.edu.sg, \{chenjingjing,ygj\}@fudan.edu.cn
}

\usepackage{bibentry}

\begin{document}

\maketitle

\begin{abstract}
Existing reinforcement learning (RL) methods struggle with long-horizon robotic manipulation tasks, particularly those involving sparse rewards. While action chunking is a promising paradigm for robotic manipulation, using RL to directly learn continuous action chunks in a stable and data-efficient manner remains a critical challenge. 
This paper introduces AC3 (Actor-Critic for Continuous Chunks), a novel RL framework that learns to generate high-dimensional, continuous action sequences. To make this learning process stable and data-efficient, AC3 incorporates targeted stabilization mechanisms for both the actor and the critic. First, to ensure reliable policy improvement, the actor is trained with an asymmetric update rule, learning exclusively from successful trajectories. Second, to enable effective value learning despite sparse rewards, the critic's update is stabilized using intra-chunk $n$-step returns and further enriched by a self-supervised module providing intrinsic rewards at anchor points aligned with each action chunk. We conducted extensive experiments on 25 tasks from the BiGym and RLBench benchmarks. 
Results show that by using only a few demonstrations and a simple model architecture, AC3 achieves superior success rates on most tasks, validating its effective design.
\end{abstract}

\begin{links}
    \link{Code}{https://github.com/flyfaerss/ac3}
\end{links}

\section{Introduction}

Recent reinforcement learning (RL) algorithms \cite{haarnoja2018soft,kalashnikov2018scalable,kalashnikov2021mt,yarats2021mastering,herzog2023deep} have demonstrated significant advances in learning continuous-action control from online experiences. Nevertheless, these methods excel mostly at short-horizon tasks such as {\it closing drawers} or {\it doors}, and consistently struggle to learn effective policies for long-horizon problems, i.e., those comprising multiple sub-tasks and requiring precise manipulation, such as {\it moving plates} or {\it sandwich flipping}.

To complete long-horizon tasks, the agent must execute an extended sequence of coherent actions, vastly expanding the state-action exploration space. In most cases, the failure of any single sub-task directly leads to failure in exploring the overall task. As a result, this learning challenge is further compounded by sparse rewards: the agent can only obtain positive feedback upon task completion, leaving exploration in intermediate steps without effective guiding signals. Consequently, it becomes exceedingly difficult for the agent to determine which preceding actions are crucial for eventual success, leading to inefficient autonomous exploration. 

\begin{figure}[t]
\begingroup
\begin{center}
\includegraphics[width=\linewidth]{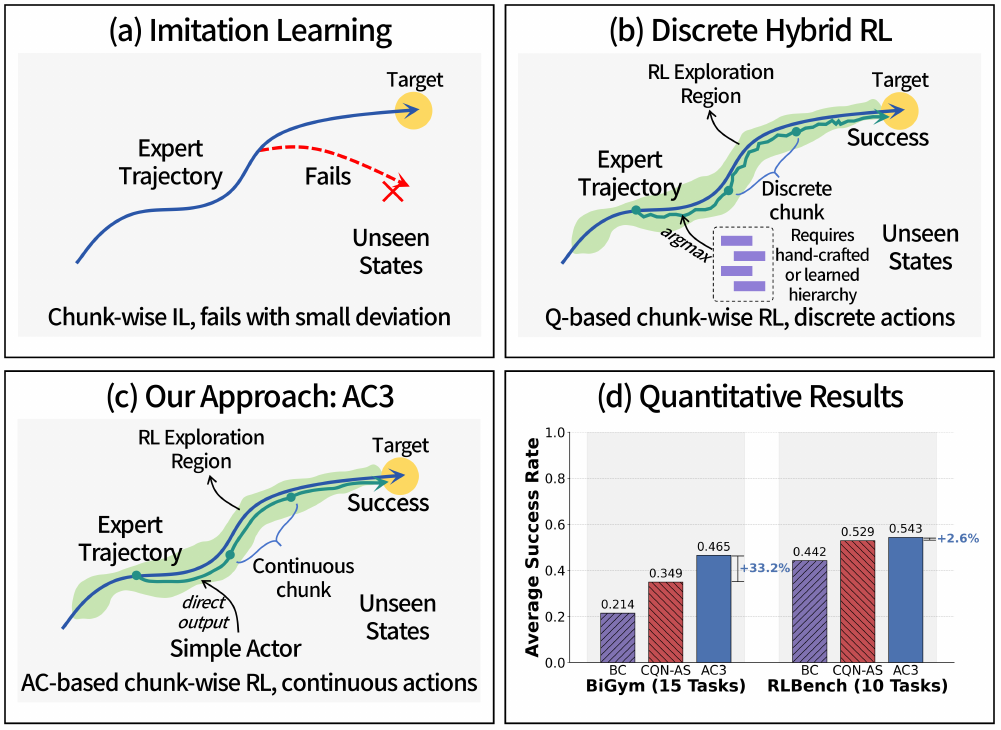}
\end{center}
   \caption{(a) Imitation Learning is not robust to unseen states. (b) Hybrid RL with discrete chunks lacks precision. (c) Our approach, AC3, directly learns continuous chunks for more effective control. (d) AC3 achieves superior performance.}
\label{fig:motivation}
\endgroup
\end{figure}

For long-horizon tasks, predicting entire sequences of actions—a paradigm known as action chunking—has proven highly successful, finding its most natural application in Imitation Learning (IL). This success is exemplified by recent methods like ACT \cite{zhao2023learning} and Diffusion Policy \cite{chi2023diffusion}, which use powerful generative models to effectively clone complex, continuous action sequences from expert data. Despite their power, these IL methods are constrained by an upper bound, as their performance is limited by the provided expert demonstrations. Due to distributional shift \cite{ross2011reduction}, the agent is likely to fail when it encounters unseen states that deviate even slightly from the expert trajectory (Fig. \ref{fig:motivation}(a)). 
The promise of overcoming these limitations, through online interaction and the potential to discover superior policies, provides a motivation to integrate the principles of RL.

However, bringing the chunking paradigm into RL is non-trivial.
The complexity of the exploration space increases exponentially as the chunk length grows, making exploration and value estimation intractable for standard methods and frequently leading to exploding $Q$-value estimates \cite{seo2025coarse} and subsequent training instability.
Recent work CQN-AS \cite{seo2025coarse} addressed this by discretizing a library of chunks and learning a $Q$-function over that set, but the discretization inevitably sacrifices precision and flexibility (as shown in Fig. \ref{fig:motivation}(b)). 
Although other concurrent work \cite{li2025reinforcement} has attempted to generate continuous action chunks via complex distillation pipelines, these methods often rely on large-scale offline datasets and are computationally expensive, limiting their application in few-shot, real-time control settings.
Consequently, a critical gap remains in the field: an efficient RL framework that can directly and data-efficiently leverage the full potential of action chunking in continuous domains.

To bridge this critical gap, this paper introduces AC3 (Actor-Critic for Continuous Chunks), a novel framework designed to efficiently learn continuous action chunks by leveraging a small number of expert demonstrations. AC3 builds directly upon a DDPG-style framework \cite{lillicrap2015continuous} to predict continuous action chunks, thereby enabling more flexible and precise robotic control. However, ensuring the stability of such a framework is paramount, especially when learning from limited data under sparse rewards.

To this end, AC3 incorporates two key innovations for stabilization. 
First, to ensure reliable policy improvement, the actor is trained with an asymmetric update rule: its policy network learns exclusively from successful trajectories (including expert demonstrations and successful online rollouts). This approach strictly confines the policy optimization to the ``trusted region'' where the value function is most reliable, thus avoiding misleading gradients from inaccurate $Q$-values and guaranteeing stable learning. 
Second, to stabilize the critic's learning under sparse rewards, its update utilizes intra-chunk $n$-step returns and is further enhanced by a self-supervised module. This module employs a goal network, pre-trained on expert demonstrations, to provide intrinsic reward signals at anchor points aligned with action chunks, thereby effectively guiding the value function's learning. Together, these components enable AC3 to robustly learn complex, continuous control policies in a data-efficient manner.

To validate the efficacy of AC3, we conducted extensive experiments on 25 robotic tasks from the BiGym \cite{chernyadev2024bigym} and RLBench \cite{james2020rlbench} benchmarks. The results show that on long-horizon, sparse-reward tasks with high-dimensional states, AC3 achieves superior success rates (Fig. \ref{fig:motivation}(d)) while utilizing a simple model architecture, thus validating its effective and data-efficient design.

The main contributions can be summarized as follows:
\begin{itemize}
    \item We propose AC3, a novel actor-critic framework for learning continuous action chunks. It is designed to tackle long-horizon, sparse-reward manipulation tasks by efficiently leveraging only a small number of expert demonstrations.
    \item We introduce two key stabilization mechanisms to ensure stable learning from sparse rewards: (1) an asymmetric update rule that trains the actor exclusively on successful trajectories for robust policy improvement, and (2) a combined critic update method leveraging intra-chunk $n$-step returns and a self-supervised module that provides chunk-wise intrinsic rewards.
    \item We conduct extensive experiments on 25 tasks from the BiGym and RLBench benchmarks, demonstrating that AC3 achieves superior success rates over existing methods by utilizing a simple model architecture, thus validating the effectiveness and data-efficiency of our approach.
\end{itemize}

\section{Related Work}
\label{sec:related_work}

\noindent{\bf Imitation Learning from Demonstrations. } Imitation Learning (IL) offers a data-driven approach to policy learning by leveraging expert demonstrations, avoiding the exploration challenges of pure RL. The most fundamental method, Behavioral Cloning (BC), treats policy learning as a supervised problem but is known to suffer from distributional shift \cite{ross2011reduction}. To better address long-horizon tasks, the IL paradigm has evolved towards action chunking, where the policy learns to predict entire sequences of actions. State-of-the-art methods in this domain, such as ACT \cite{zhao2023learning,lee2024interact} and Diffusion Policy \cite{chi2023diffusion,ren2024diffusion,wang2025hierarchical}, employ powerful generative models (e.g., Transformers \cite{vaswani2017attention} and diffusion models \cite{ho2020denoising}) to approximate the action distribution, and have been extended to multi-task setups \cite{bharadhwaj2024roboagent,liu2024rdt,doshi2024scaling}, mobile manipulation \cite{fu2024mobile,motoda2025learning} and humanoid control \cite{fu2024humanplus}.
However, their performance is fundamentally capped by the expert data and they cannot improve through online interaction, which motivates integrating the principles of RL.

\noindent{\bf Reinforcement Learning with Action Chunks. } Structured modeling of action sequences is an important approach to improving the efficiency of policy learning and execution \cite{saanum2023reinforcement}. However, applying the action chunking paradigm to a RL context is non-trivial, as it dramatically expands the action space for an exploring agent. Discovering an effective policy through exploration in long-horizon, sparse-reward environments is quite difficult, thus existing works generally rely on demonstrations for exploration guidance. ResiP \cite{ankile2024imitation} applies RL as a residual component to fine-tune action chunk predicted by BC, but its RL process still performs policy updates at the low-level, single-timestep scale. 
T-SAC \cite{tian2025chunking} learns a critic on action chunks by integrating $n$-step returns while still optimizing a single-timestep actor.
CQN-AS \cite{seo2025coarse}, in contrast, uses the action chunk as the high-level unit for RL exploration. However, its reliance on $Q$-learning forces a discretization of the action space, fundamentally limiting the policy's precision in continuous control. While more recent work $Q$-Chunking \cite{li2025reinforcement} overcomes this by generating continuous chunks via a complex distillation pipeline, it introduces computational overhead and 
relies on large-scale offline datasets. 
Therefore, developing a framework that can learn continuous action chunks in a direct, stable, and computationally efficient manner remains a challenge. To address this challenge, we introduce AC3.

\section{Method}

\begin{figure}[t]
\begingroup
\begin{center}
\includegraphics[width=\linewidth]{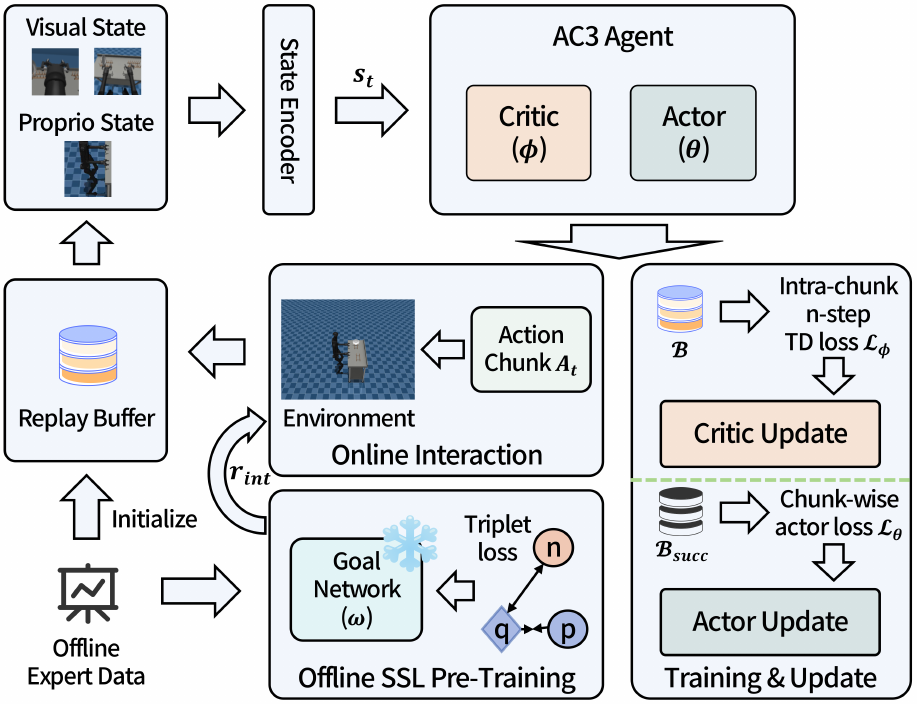}
\end{center}
   \caption{Overall framework of AC3. First, a Goal Network is pre-trained using expert data via self-supervised learning to provide intrinsic rewards $r_{\text{int}}$ during subsequent online interactions. Next, during online interaction, the Actor outputs a continuous action chunk and stores new experiences in the Replay Buffer after execution. For training, the Critic is updated via an intra-chunk $n$-step TD loss, while the Actor learns only from the successful trajectories buffer $\mathcal{B}_{\text{succ}}$ to promote stable policy improvement.}%
\label{fig:framework}
\endgroup
\end{figure}

Our goal is to tackle long-horizon robotic manipulation tasks where reward is granted only upon success, using just a few expert demonstrations—a setting that naturally reflects real-world data constraints.
To this end, we introduce AC3 (Actor-Critic for Continuous Chunks), an off-policy actor-critic framework designed to learn a policy that generates sequences of continuous actions. 
As illustrated in Fig. \ref{fig:framework}, its core design follows three levels:
First, we construct a basic AC3 Agent, whose Actor network is designed to directly output continuous action chunks, while the Critic network is responsible for evaluating the value of these chunks. Second, in terms of update rules, we adopt intra-chunk $n$-step returns to stabilize the Critic's training, while constraining the Actor's update to use only transitions from successful trajectories to ensure reliable policy optimization. Third, we introduce a self-supervised reward shaping module, which sets anchor points in units of action chunks to provide relatively dense and reliable intrinsic rewards for the Critic, thereby effectively guiding the learning of the value function.

\subsection{Preliminaries: Problem Formulation}

We model the robotic manipulation task as a semi-Markov Decision Process (SMDP) \cite{sutton1999between} with continuous chunk space, defined by the tuple $(\mathcal{S}, \mathcal{A}, P, r, \gamma)$. Here, $\mathcal{S}$ is a high-dimensional state space, which includes RGB visual states and the robot's proprioceptive states. $\mathcal{A}$ is a continuous action chunk with fixed length, $P$ is a state transition dynamics, $\gamma \in [0, 1)$ is a discount factor, and $r$ is the reward function. For the tasks we consider, the reward $r \in \{0, 1\}$ is sparse, providing a positive signal only upon successful task completion. We utilize a small offline dataset of $N$ expert demonstrations $\mathcal{D}_{\text{demo}} = \{\tau_{1}, ..., \tau_{N}\}$, where each trajectory $\tau$ is a sequence of state-action pairs. Our objective is to learn a policy $\pi_{\theta}(s)$ that maximizes the expected discounted return $\mathbb{E}[\sum_{t=0}^{T} \gamma^{t}r_{t}]$.

\subsection{AC3 Algorithm}

AC3 is an off-policy actor-critic algorithm derived from the DDPG-style framework. It is specially designed for long-horizon tasks by making two principal modifications: 1) the actor network outputs temporal action chunks; 2) the critic network is trained using intra-chunk $n$-step returns. The training pipeline of AC3 is summarized in Algorithm \ref{alg:algorithm}.

\noindent{\bf Actor: Continuous Action Chunking Policy. } Temporal abstraction is a critical technique in long-horizon robotic manipulation tasks. It aligns motion planning with the core temporal logic of the entire task, thereby mitigating local sub-optimalities that arise from step-wise decision-making. At the same time, this approach also enhances the control over the long-horizon operational sequence, which in turn improves the coherence and precision of action execution.
Based on this principle, our policy $\pi_{\theta}$, parameterized by $\theta$, directly predicts a continuous action chunk $\mathcal{A}_{t} \in \mathbb{R}^{C\times d_{a}}$, which is a sequence of $C$ continuous actions:
\begin{equation}
\mathcal{A}_{t} = \pi_{\theta}(s_t) = \{a_{t}, a_{t+1}, ..., a_{t+C-1}\},
\end{equation}
where $s_t$ is the joint state representation at timestep $t$, which is formed by concatenating the robot's proprioceptive state features and current RGB visual features. 
To capture the temporal dependencies within action sequences while ensuring computational efficiency, the policy network employs a simple MLP+GRU architecture.

\noindent{\bf Critic: Intra-Chunk $n$-step $Q$-value Estimation. } The critic, $Q_{\phi}(s_t, \mathcal{A}_t)$ with parameters $\phi$, estimates the expected return after executing the entire chunk $\mathcal{A}_t$ starting from state $s_t$. 
In long-horizon tasks, a relatively large chunk size ({\it e.g.}, 16) is crucial for capturing temporal dependencies and also demonstrates significant performance gains in complex tasks (see Fig. \ref{fig:action_chunk_length}). However, in such cases, directly evaluating the $Q$-value of the entire chunk leads to two key issues: 1) The reward estimation is more noisy, resulting in reduced sample efficiency. As a result, more diverse data are required to converge to an effective policy, yet robot data collection is inherently challenging; 2) The Critic cannot provide fine-grained feedback for each primitive action within the chunk, thereby losing the precision in action execution.

To improve training stability and sample efficiency, we instead use $n$-step TD target to update the critic. This target is composed of the observed reward sum from the replay buffer and a bootstrapped value estimate from the target networks $(Q_{\phi'})$ at the landing state $s_{t+n}$. The target value is formulated as:
\begin{equation}
y_t = \sum_{k=0}^{n-1} \gamma^{k} r_{t+k} + \gamma^{n}\mathop{\min}_{i=1,2}Q_{\phi'_{i}}(s_{t+n}, \pi_{\theta}(s_{t+n})+\epsilon),
\label{eq:target_value}
\end{equation}
similar to TD3 \cite{fujimoto2018addressing}, the exploration noise $\epsilon$ is sampled from $\text{clip}(\mathcal{N}(0, \sigma), -c, c)$, and we keep two target critics and use their clipped double-Q estimate $\min_{i} Q_{\phi'_{i}}$ to curb over-estimation bias and stabilize learning. The critic network still employs a simple MLP+GRU architecture similar to actor network. 

\noindent{\bf Training and Updates. } The actor and critic are trained by sampling mini-batches of transitions $\tau = (s_t, \mathcal{A}_{t}^{\text{exp}}, r_{t:t+n-1}, s_{t+n})$ from the replay buffer $\mathcal{B}$.

\noindent{\it Critic Update: } The critic's role is to learn an accurate and robust value function by minimizing the TD error, thus providing global guidance for the policy gradient update direction. Specifically, the parameters $\phi$ of both critic networks are updated by minimizing the MSE loss between the current $Q$-value predictions and the target value $y_{t}$ calculated in Eq. \ref{eq:target_value}. The combined critic loss is:
\begin{equation}
\mathcal{L}_{\phi} = \sum_{i=1,2}\mathbb{E}_{\tau \sim \mathcal{B}} \left[(Q_{\phi_i}(s_t, \mathcal{A}_{t}^{\text{exp}}) - y_t)^2 \right].
\end{equation}

\noindent{\it Actor Update: } Under the conditions of high dimensionality of the state space and sparse rewards, the blind exploration of RL will become inefficient. Therefore, a common practice is to use expert demonstrations for guidance. Specifically, we construct a new replay buffer $\mathcal{B}_{\text{succ}}$, which filters the successful trajectory transitions from $\mathcal{B}$. After that, we directly use the BC loss for imitation.
\begin{equation}
\mathcal{L}_{\theta_{BC}} = \mathbb{E}_{(s_t, \mathcal{A}_{t}^{\text{exp}}) \sim \mathcal{B}_{\text{succ}}}\left[ \Vert \pi_{\theta}(s_t) - \mathcal{A}_{t}^{\text{exp}} \Vert^{2} \right].
\end{equation}
To ensure the policy actively improves itself by leveraging online experience, we further incorporate the $Q$-value signal provided by the critic:
\begin{equation}
\mathcal{L}_{\theta_{Q}} = -\mathbb{E}_{s_t \sim \mathcal{B}_{\text{succ}}}\left[ \mathop{min}_{i=1,2} Q_{\phi_i}(s_t, \pi_{\theta}(s_t)) \right].
\end{equation}
Unlike the critic network, we found that if the actor is trained on all transitions, the model's performance actually degrades (see Fig. \ref{fig:training_stability}). This phenomenon is intuitive: in a high-dimensional state space and under conditions of extremely sparse rewards, the critic cannot assign an accurate $Q$-value to most of the state space, and a new state might cause the model to update in an unpredictable direction. Therefore, we strictly constrain the calculation of the policy gradient to the ``trusted region'' defined by the successful trajectories. In this way, the policy's optimization is confined to the regions where the value function is most reliable, effectively avoiding the policy degradation that could result from exploring uncertain states.

The final actor loss is a weighted combination of the BC loss and the Q loss:
\begin{equation}
\mathcal{L}_{\theta} = \lambda_{BC} \mathcal{L}_{\theta_{BC}} + \lambda_{Q} \mathcal{L}_{\theta_{Q}},
\label{eq:total_loss}
\end{equation}
where $\lambda_{BC}$ and $\lambda_{Q}$ are weighting coefficients.

\begin{algorithm}[tb]
    \caption{AC3}
    \label{alg:algorithm}
    Initialize critic networks $Q_{\phi_1}$, $Q_{\phi_2}$, and actor network $\pi_\theta$ with random parameters $\phi_1$, $\phi_2$, $\theta$ \\
    Initialize critic target networks $\phi'_{1}\leftarrow \phi_1$, $\phi'_{2} \leftarrow \phi_2$ \\
    Initialize replay buffer $\mathcal{B}$ with expert demonstrations $\mathcal{D}_{\text{demo}}$ \\
    Pre-train goal network $G_{\omega}$ with parameters $\omega$ using Eq. \ref{eq:triplet_loss} \\
    \hspace*{-1.2\leftmargin}
    \parbox[l]{\dimexpr\linewidth-\algorithmicindent}{
    \begin{algorithmic}
        \For{timestep $t = 1 \text{ to } T$} 
            \If{$t \text{ mod } C$}
                \State \parbox[t]{\dimexpr\linewidth-\algorithmicindent}{Compute action chunk with exploration noise $\mathcal{A}_{t}^{\text{exp}} \sim \pi_{\theta}(s_t) + \epsilon$, $\epsilon \sim \text{clip}(\mathcal{N}(0, \sigma), -c, c)$}
            \EndIf
            \State \parbox[t]{\dimexpr\linewidth-\algorithmicindent + 1.2\leftmargin}{Pop action $a_t$ from current action chunk, execute $a_t$ in the environment, obtain reward $r_{t}$ and new state $s_{t+1}$}
            \State $r_t \leftarrow r_{\text{int}}(s_t)$ using Eq. \ref{eq:intrinsic_reward} if not terminal $s_t$
            \State Store transition tuple $(s_t, a_t, r_t, s_{t+1})$ in $\mathcal{B}$
            
            \Statex
            \State \parbox[t]{\dimexpr\linewidth-\algorithmicindent + 1.2\leftmargin}{Sample $B$ transitions $(s_t, \mathcal{A}_{t}^{\text{exp}}, r_{t:t+n-1}, s_{t+n})$ from replay buffer $\mathcal{B}$ as a mini-batch}
            \State \textit{\textbf{// Critic Update}}
            \State Compute target value $y_t$ using Eq. \ref{eq:target_value}
            \State $\mathcal{L}_{\phi_i} = \frac{1}{B}\sum[(Q_{\phi_i}(s_t, A_{t}^{\text{exp}}) - y_t)^{2}]~~~~\forall i \in \{1,2\}$
            \State $\phi_i \leftarrow \phi_i - \alpha \nabla_{\phi_i} \mathcal{L}_{\phi_i}~~~~\forall i \in \{1,2\}$
            \State $\phi'_i \leftarrow (1-\mu)\phi'_i + \mu\phi_i~~~~\forall i \in \{1,2\}$

            \State \textit{\textbf{// Actor Update}}
            \State \parbox[t]{\dimexpr\linewidth-\algorithmicindent+1.2\leftmargin}{Filter $B_{\text{succ}}$ successful trajectories' transitions from current mini-batch through successful mask $m_{\text{succ}}$}
            \State $\mathcal{L}_{\theta_{BC}} = \frac{1}{B_{\text{succ}}} \sum \left[\Vert \pi_{\theta}(s_t) - \mathcal{A}_{t}^{\text{exp}} \Vert^{2} * m_{\text{succ}} \right]$
            \State $\mathcal{L}_{\theta_{Q}} = - \frac{1}{B_{\text{succ}}} \sum\left[ \min_{i=1,2} Q_{\phi_i}(s_t, \pi_{\theta}(s_t)) * m_{\text{succ}} \right]$
            \State $\mathcal{L}_{\theta} = \lambda_{BC} \mathcal{L}_{\theta_{BC}} + \lambda_{Q} \mathcal{L}_{\theta_{Q}}$
            \State $\theta \leftarrow \theta - \alpha \nabla_{\theta} \mathcal{L}_{\theta}$
        \EndFor
    \end{algorithmic}}
\end{algorithm}

\subsection{Self-Supervised Reward Shaping}

Eq. \ref{eq:total_loss} utilizes a few offline demonstrations to provide direct guidance for RL exploration. However, the environment remains reward-sparse, thus the critic network still struggles to learn a meaningful state-value function. To alleviate this issue, we further construct a self-supervised reward shaping module by pre-training a state goal network, $G_{\omega}(\cdot)$, on the offline demonstrations $\mathcal{D}_{\text{demo}}$, with the same architecture as the state encoder. The goal of $G_{\omega}$ is to learn a low-dimensional latent space that captures the key features of successful trajectories.

\begin{figure*}[htbp]
\begin{center}
\includegraphics[width=0.95\linewidth]{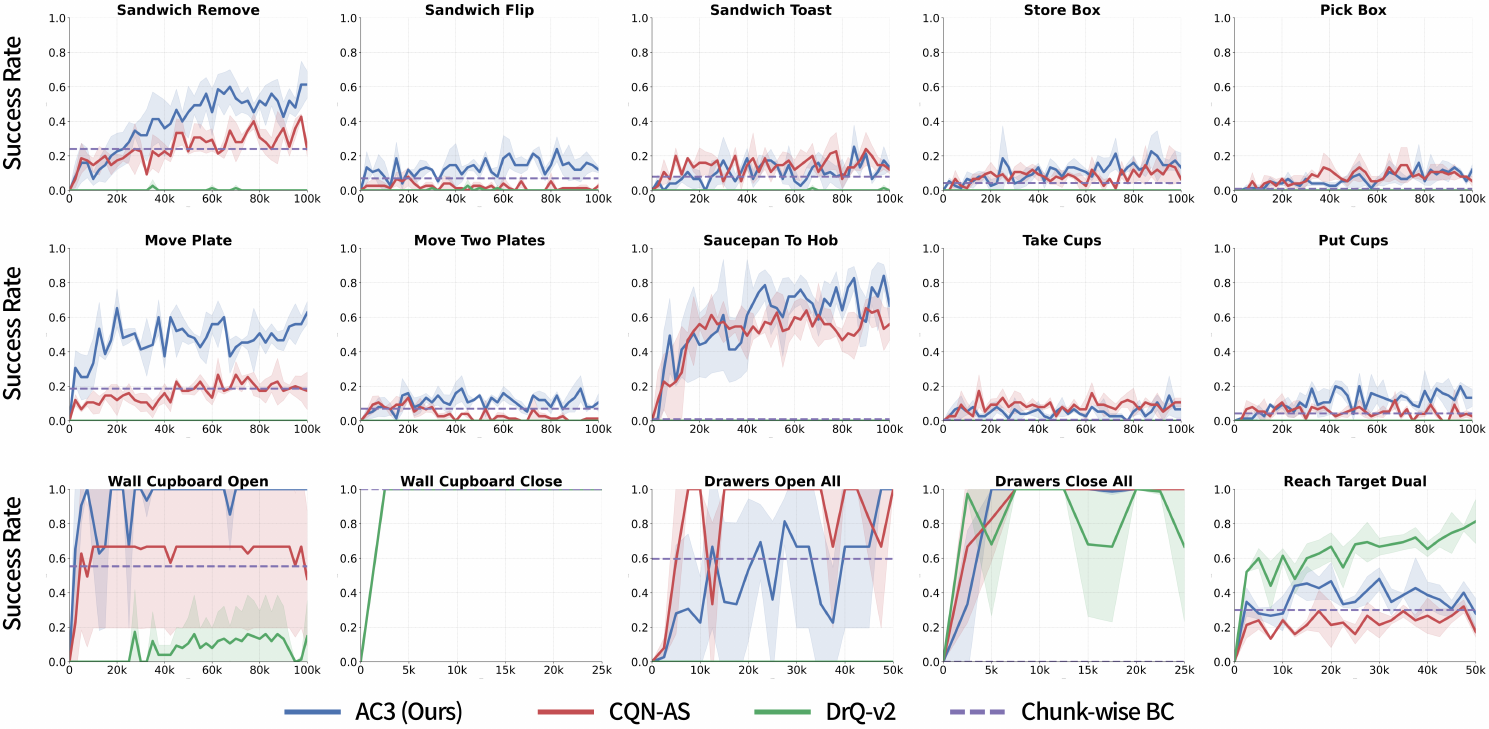}
\end{center}
\caption{The performance of 15 bi-manual mobile manipulation tasks in BiGym. All tasks use 10 expert demonstrations as offline data, and all RL algorithms use an auxiliary BC loss for exploration guidance. The solid line and the shaded regions represent the mean performance and standard deviation, respectively.}
\label{fig:bigym_main}
\end{figure*}

Specifically, we employ a contrastive learning paradigm to train $G_{\omega}$. States that are temporally close within the same demonstration trajectory form positive pairs, while states from different trajectories or those temporally distant in the same trajectory form negative pairs. Triplet loss is used to pull positive pairs closer together and push negative pairs further apart in the embedding space.
\begin{equation}
\begin{aligned}
\mathcal{L}_{\omega} =& \mathbb{E}_{(s_q, s_p, s_n) \sim \mathcal{D}_{\text{demo}}} [ \max ( \| G_\omega(s_q) - G_\omega(s_p) \|^2 \\
&- \| G_\omega(s_q) - G_\omega(s_n) \|^2 + m, 0 ) ],
\end{aligned}
\label{eq:triplet_loss}
\end{equation}
where $s_q$ is the query sample, $s_p$ is the positive sample, $s_n$ is the negative sample, and $m$ is a margin hyperparameter that enforces the distance to the negative sample to be greater than the distance to the positive sample by at least $m$.

During the online policy learning, we introduce a semi-dense reward mechanism based on ``Anchor Points''. Instead of calculating an intrinsic reward at every timestep, we set Anchor Points at a fixed interval of $K$ timesteps ({\it e.g.}, $K=C$) along the exploration trajectory. The intrinsic reward calculation is activated only when a timestep $t$ is an Anchor Point ({\it i.e.}, $t \pmod{K} = 0$). This design naturally aligns with our chunk-based AC3 policy, effectively providing an intrinsic reward signal to the entire action chunk and encouraging random exploration within it. 

To formalize the reward calculation, we first define the minimum squared latent distance, $d(s_t)$, between the current state $s_t$ and the set of demonstration states:
\begin{equation}
d(s_t) = \min_{s_{d}\in \mathcal{D}_{\text{demo}}}\Vert G_{\omega}(s_t) - G_{\omega}(s_{d}) \Vert^{2}.
\label{eq:latent_distance}
\end{equation}
Using this distance metric, the intrinsic reward $r_{\text{int}}(s_t)$ is then given by:
\begin{equation}
r_{\text{int}}(s_t) =
\begin{cases}
a  &\text{if } t \equiv 0 ~~(\bmod \text{ }K) ~~\text{and}~~ d(s_t) < m, \\
0  &\text{otherwise.}
\end{cases}
\label{eq:intrinsic_reward}
\end{equation}
Here, $m$ is equivalent to the margin threshold in Eq. \ref{eq:triplet_loss}, thus ensuring the anchor state is always near the successful trajectory. In addition, $a$ can be any positive reward that is not greater than the task success reward, thereby functioning as an exploration anchor. We simply set 0.1 in our experiments.

\section{Experiments}

\subsection{Experimental Setup}

\noindent{\bf Dataset and Setup. } 
We conduct evaluations on two robotic manipulation benchmarks with sparse reward.

\begin{itemize}
    \item {\bf BiGym}: BiGym \cite{chernyadev2024bigym} is a benchmark focused on bi-manual mobile manipulation tasks. Each task contains human-collected demonstrations, with each trajectory receiving a reward of 1 only upon task success, and 0 otherwise. We select 15 tasks covering scenarios like multiple sub-tasks ({\it e.g.}, {\it saucepan to hob}) and fine-grained operations ({\it e.g.}, {\it move plate}). We use RGB observations with 84×84 resolution from {\it head}, {\it left wrist}, and {\it right wrist} cameras. For each task, we select 10 demonstrations as offline expert data in each run.

    \item {\bf RLBench}: RLBench \cite{james2020rlbench} is a benchmark focusing on tabletop manipulation. We select 10 tasks, with the same reward setting as BiGym. We use RGB observations with 84×84 resolution from {\it front}, {\it wrist}, {\it left shoulder}, and {\it right shoulder} cameras. We maintain the same demonstration collection method as CQN \cite{seo2024continuous}, with 100 demonstrations per task.
\end{itemize}

\noindent{\bf Baseline and Evaluation Metrics. }
We report the performance of four algorithms: 1) Our proposed AC3; 2) CQN-AS \cite{seo2025coarse}, which learns action chunks using hierarchical discrete $Q$-learning; 3) DrQ-v2 \cite{yarats2021mastering}, an actor-critic-based deterministic policy algorithm that outputs actions for each timestep. To enable DrQ-v2 to converge to effective policies in sparse reward settings, we employ an additional auxiliary BC loss; 4) Chunk-wise BC, which uses the same policy ({\it i.e.}, Actor) model architecture of AC3 and only employs offline data for chunk-based imitation learning. We train 50k steps on each task, and take the mean performance of the last three checkpoints as the chunk-wise BC baseline, the quantitative results shown in Fig. \ref{fig:motivation}(d) are also calculated using a similar rule. 
We report the success rate of 25 episodes at each checkpoint. In all figures we plot the mean performance over 3 seeds together with the shaded regions representing the standard deviation.

\begin{figure*}[htbp] 
\begin{center}
\includegraphics[width=0.95\linewidth]{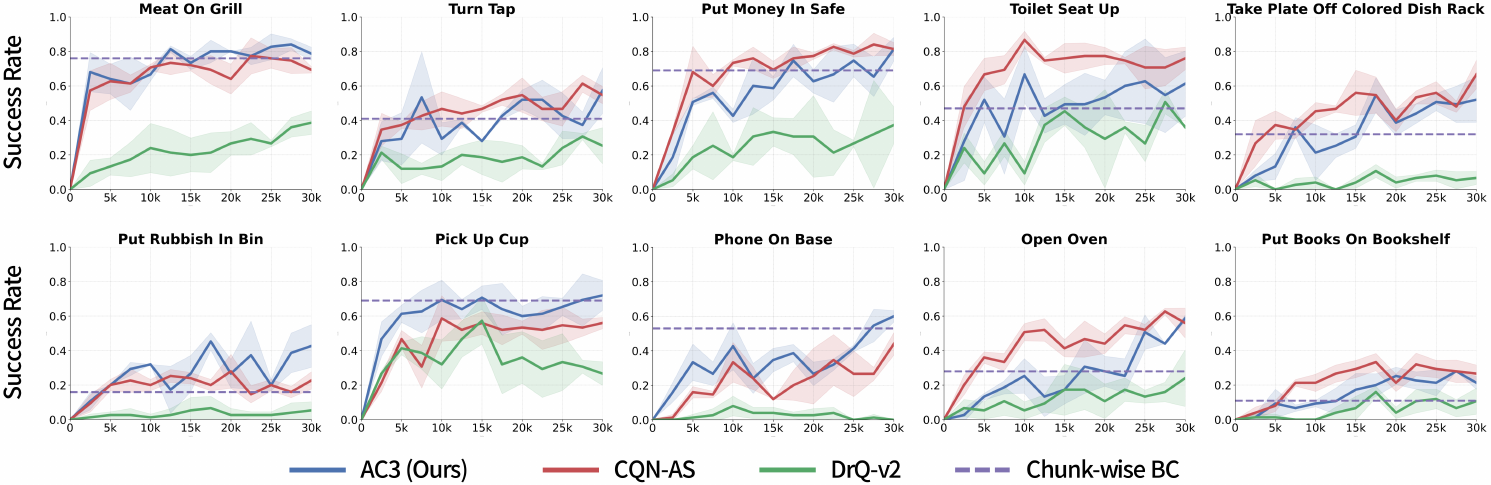}
\end{center}
\caption{The performance of 10 tabletop manipulation tasks in RLBench. All tasks use 100 synthetic demonstrations as offline data, and all RL algorithms use an auxiliary BC loss for exploration guidance. The solid line and the shaded regions represent the mean performance and standard deviation, respectively.}
\label{fig:rlbench_main}
\end{figure*}

\noindent{\bf Implementation Details.}  
We use the AdamW optimizer for all networks. The learning rates are set to $1.0\times 10^{-4}$ for the actor $\pi_{\theta}$ and critic $Q_{\phi}$, and $1.0\times 10^{-5}$ for the goal network $G_{\omega}$. 
For the basic RL settings, we set soft update coefficient for the target networks $\mu=0.005$, discount factor $\gamma=0.99$, exploration noise $\sigma=0.01$, and its clip threshold $c=0.1$. We employ the temporal ensemble method proposed in ACT to smooth action execution. For intrinsic reward shaping, we set anchor interval $K=16$, margin threshold $m=0.5$, and intrinsic reward $a=0.1$. In the actor loss function, $\lambda_{BC} = 1.0$ and $\lambda_{Q} = 0.1$. 
In the main results, we set the action chunk size $C = 16$. For network updates, BiGym and RLBench tasks both use 4-step returns. All our experiments are conducted using a RTX 3090 GPU.

\subsection{Performance Comparison}

\noindent{\bf BiGym Results. }
As shown in Fig. \ref{fig:bigym_main}, with the same few demonstrations (10 demos), AC3 achieves the highest task success rate in most tasks. Compared with the BC baseline, AC3 consistently discovers a better strategy and rapidly increases the success rate within 10k steps. For instance, in tasks like {\it sandwich remove} and {\it move plate}, its success rate curve exhibits a more prominent upward trend and maintains a relatively high level, indicating the effectiveness and stability of AC3 in exploring the high-dimensional action.

\noindent{\bf RLBench Results. }
The tasks in RLBench not only provide cleaner and more demonstrations but also feature smaller action spaces. As a result, as shown in Fig. \ref{fig:rlbench_main}, chunk-wise BC can typically learn effective policies for task completion. Building on this, AC3 still outperforms the BC baseline on most tasks and, with a simpler network architecture, achieves similar performance to CQN-AS. In contrast, DrQ-v2 performs significantly worse than the BC baseline on most tasks, demonstrating the critical role and necessity of chunk-based prediction in long-horizon, sparse-reward tasks.

\begin{figure}[tbp]
\begin{center}
\includegraphics[width=0.95\linewidth]{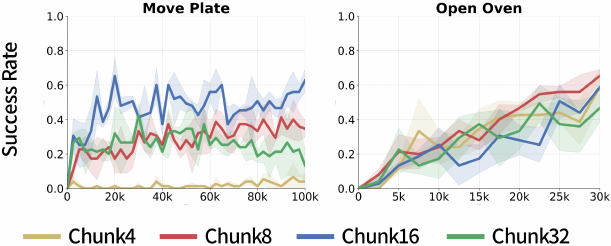}
\end{center}
\caption{Effect of action chunk length.}
\label{fig:action_chunk_length}
\end{figure}

\subsection{Ablation Study and Analysis}
In the ablation study, we primarily explore several key designs of AC3, with our experiments focusing mainly on comparing the representative {\it move plate} task from BiGym and {\it open oven} task from RLBench: 1) {\it How the size of action chunks affects policy}; 2) {\it Whether the intra-chunk $n$-step return impacts policy stability}; 3) {\it Whether the self-supervised reward design facilitates policy training}; 4) {\it Training stability and efficiency analysis}.

\noindent{\bf Discussion on action chunk length. }
The length of action chunks $C$ is crucial for AC3 training. Larger action chunks provide longer action consistency and incorporate more temporal dependency information, but they also increase the dimensionality of the action exploration space and introduce more exploration noise. As shown in Fig. \ref{fig:action_chunk_length}, in the {\it move plate} task with high-dimensional action space, a longer action chunk ensures the coherence and effectiveness of the bi-manual movements, whereas a setting of $C=4$ renders the policy almost completely ineffective. In contrast, for the simpler {\it open oven} task with fewer DoF, overly long action chunks increase exploration difficulty and noise, while a moderate chunk length is sufficient to strike the optimal balance between action coherence and exploration efficiency.

\begin{figure}[tbp]
\begin{center}
\includegraphics[width=0.95\linewidth]{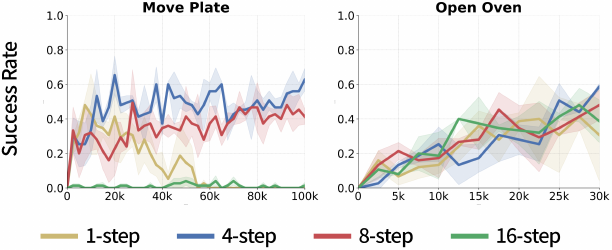}
\end{center}
\caption{Effect of intra-chunk $n$-step return.}
\label{fig:nstep}
\end{figure}

\noindent{\bf Effectiveness of intra-chunk $n$-step return. }
In theory, $n=C$ is an unbiased choice under ideal conditions. However, this approach reduces the policy's prediction units to entire chunks: larger chunks introduce greater exploration noise, and the critic fails to provide precise guidance for intra-chunk actions. In environments with limited demonstrations and sparse rewards, such noise and variance can be critical. As shown in Fig. \ref{fig:nstep}, $n=C$ ({\it i.e.}, 16) setting in the {\it move plate} task directly leads to policy failure, whereas performance is significantly better in {\it open oven} task—where offline data is more diverse and action space is relatively low.

Intra-chunk $n$-step return serves as a compromise: it allows minor fine-tuning within chunks, reducing exploration noise. Intuitively, the critic evaluates short-term execution against a longer-term plan, enabling the model to better balance macro planning ({\it i.e.}, which chunk to execute) and micro control ({\it i.e.}, how to optimize intra-chunk actions). By leveraging the more stable learning signals from intra-chunk $n$-step returns, it effectively eases the learning difficulty under sparse rewards, thus achieving more efficient policy improvement. We find that $n=4$ and $n=8$ are both good choices, while $n=1$ tends to cause $Q$-value explosion in higher-dimensional tasks, leading to policy failure in the later stages of training.

\begin{figure}[tbp] 
\begin{center}
\includegraphics[width=0.95\linewidth]{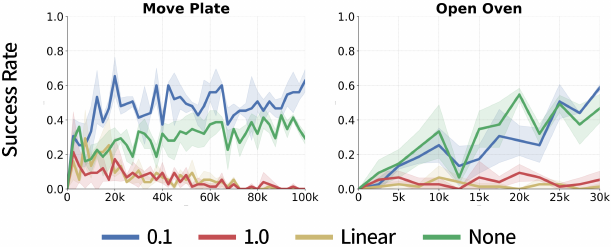}
\end{center}
\caption{Reward shaping effects. {\it Linear}: Increases linearly with the current trajectory length; {\it None}: no reward shaping.}
\label{fig:ssl_reward_setup}
\end{figure}

\noindent{\bf Analysis of self-supervised reward shaping setup. }
Setting intrinsic rewards for long-horizon, sparse-reward tasks is a common approach. We directly leverage a small amount of offline data for feature pre-training to generate anchor signals from successful trajectories, which in turn guide exploration. Fig. \ref{fig:ssl_reward_setup} shows the results for different reward settings. We find that compared to the baseline without reward shaping, $r_{\text{int}}=0.1$ exhibits a significant performance advantage in the more complex {\it move plate} task, while also maintaining decent performance in simpler {\it open oven} tasks. This demonstrates that appropriate anchor point settings play a positive role in exploration. However, we also find that when the anchor signal is either identical to the task success reward, or increases linearly as the trajectory progresses, both cases lead to $Q$-value explosion and policy failure. We attribute this to the agent learning to exploit the misspecified intrinsic reward rather than pursuing the true sparse-reward objective.

\begin{figure}[tbp]
\begin{center}
\includegraphics[width=0.95\linewidth]{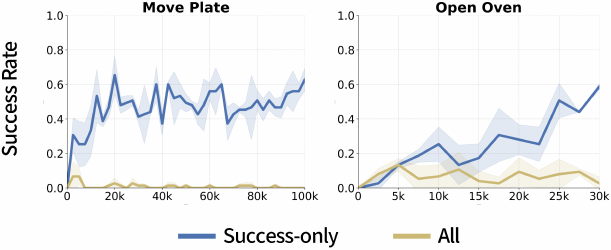}
\end{center}
\caption{Comparison between actor learning from success-only and all trajectories.}
\label{fig:training_stability}
\end{figure}

\noindent{\bf Training Stability and Efficiency Analysis. }
To demonstrate the necessity of our asymmetric update rule, we conducted an ablation study allowing the actor to learn from all experiences, including failures. As shown in Fig. \ref{fig:training_stability}, this change caused the policy to degrade dramatically or collapse entirely. This failure occurs because, in high-dimensional, sparse-reward settings, the critic generates severely biased $Q$-value estimates for unseen states. These flawed estimates produce misleading gradients that disrupt the actor's update. Therefore, AC3's asymmetric rule, which shields the actor from these noisy signals, is crucial for stable training.

Furthermore, Tab. \ref{tab:algorithm_comparison} highlights AC3's efficiency. By using only its lightweight Actor (6.56M) for inference, AC3 requires just 2.9ms to predict an action chunk, making it over 3x faster than CQN-AS. This speed is vital for high-frequency environments, as it can accommodate the overhead of action smoothing like temporal ensembling. Ultimately, this makes AC3 not only robust but also a practical choice for low-latency deployment.

\begin{table}[tbp]
\centering
\begin{tabular}{lcc}
\toprule
\textbf{Methods} & \makecell[c]{\textbf{Trainable}\\\textbf{Parameters (M)}} & \makecell[c]{\textbf{Inference}\\\textbf{Speed (ms)}} \\
\midrule
Chunk-wise BC & 6.56 (Actor-only) & 2.9 \\
CQN-AS & 28.58 (Q-Network) & 9.5 \\
AC3 & 14.44 (Actor-Critic) & 2.9 \\
\bottomrule
\end{tabular}
\caption{Comparison of model complexity and inference speed. For AC3, inference is performed using only the Actor network (6.56M parameters).}
\label{tab:algorithm_comparison}
\end{table}

\section{Conclusion}

This paper introduces AC3 (Actor-Critic for Continuous Chunks), a novel framework designed to address long-horizon, sparse-reward robotic manipulation. AC3 directly learns to generate continuous action chunks, which enables more precise and flexible control. Its stability and data efficiency originate from two key stabilization mechanisms. First, an asymmetric actor update rule ensures reliable policy improvement by learning exclusively from successful trajectories. Second, the critic's learning is stabilized with intra-chunk $n$-step returns and guided by a self-supervised intrinsic reward, which facilitates effective learning despite sparse rewards.
We evaluate AC3 on 25 tasks from the BiGym and RLBench benchmarks. The results show that AC3 achieves superior success rates with only a simple model architecture and a few expert demonstrations. By effectively leveraging online self-improvement, AC3 presents a stable and data-efficient solution for complex manipulation tasks.

\bibliography{main2026}

\appendix

\section{Appendix A. Details of Model Architecture and Hyperparameter}

\begin{figure*}[htbp] 
    \centering 

    \begin{subfigure}[b]{0.3\textwidth}
        \centering
        \includegraphics[width=0.8\textwidth]{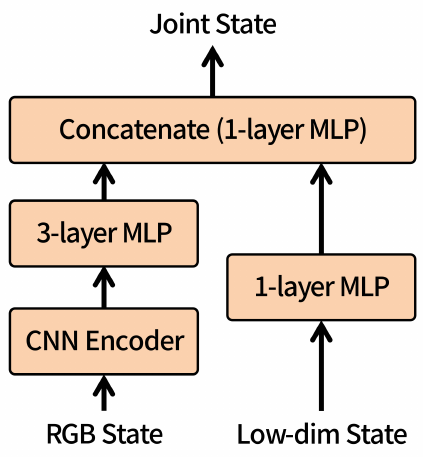} 
        \caption{Unit of state encoder.}
        \label{fig:state_encoder}
    \end{subfigure}
    \begin{subfigure}[b]{0.3\textwidth}
        \centering
        \includegraphics[width=0.8\textwidth]{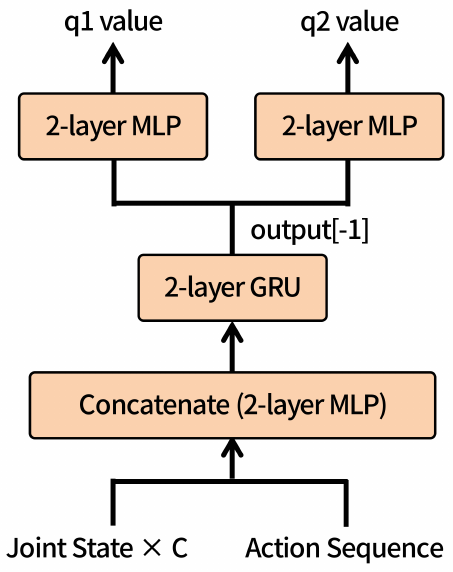} 
        \caption{Unit of critic network.}
        \label{fig:critic_network}
    \end{subfigure}
    \begin{subfigure}[b]{0.3\textwidth}
        \centering
        \includegraphics[width=0.8\textwidth]{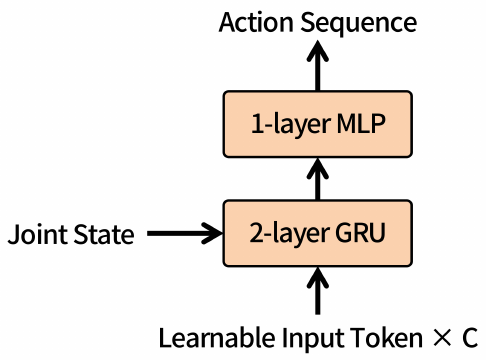}
        \caption{Unit of actor network.}
        \label{fig:actor_network}
    \end{subfigure}

    \caption{Model architecture of core modules in AC3.}
    \label{fig:model_architecture}
\end{figure*}

\begin{table*}[tbp]
\centering
\begin{tabular}{lll}
\toprule
\textbf{Module} & \textbf{Hyperparameter} & \textbf{Value} \\
\midrule
\multirow{3}{*}{\makecell[c]{$\textit{Input Image}$}} & Resolution & 84$\times$84$\times$3 \\
 & Augmentation & RandomShift \cite{yarats2021mastering}  \\
 & Frame Stack & 4 (Bigym) / 8 (RLBench) \\
\midrule
\multirow{10}{*}{\makecell[c]{$\textit{Goal Network}$}} & Architecture & Figure \ref{fig:model_architecture}(a) \\
 & Learning rate & $10^{-5}$ \\
 & Pre-training steps & 5 epochs over 5 seeds ({\it i.e.}, total 25 epochs) \\
 & Positive window & 5 timesteps \\
 & Negative margin & 0.1 * $\text{len}(\tau)$ \\
 & Output feature dim & 1024 \\
 & Loss type & Triplet loss \\ 
 & Margin threshold ($m$) & 0.5 \\
 & Anchor interval ($K$) & 16 \\ 
 & Intrinsic reward ($a$) & 0.1 \\ 
\midrule
\multirow{5}{*}{\makecell[c]{$\textit{Critic Network}$}} & Architecture & Figure \ref{fig:model_architecture}(b) \\
 & Hidden dimension & 512 \\
 & Target critic update ratio ($\mu$) & 0.005 \\
 & $n$-step return & 4 \\
 & Discount factor & 0.99 \\
\midrule
\multirow{4}{*}{\makecell[c]{$\textit{Actor Network}$}} & Architecture & Figure \ref{fig:model_architecture}(c) \\
 & Hidden dimension & 512 \\
 & BC loss coefficient ($\lambda_{BC}$) & 1.0 \\
 & RL loss coefficient ($\lambda_{Q}$) & 0.1 \\
\midrule
\multirow{7}{*}{\makecell[c]{$\textit{Other Training Setting}$}} & Action chunk size ($C$) & 16 \\
 & Batch size & 256 \\
 & Optimizer & AdamW \\
 & Learning rate & $10^{-4}$ \\
 & Weight decay & 0.01 \\
 & Action mode & Absolute Joint (for Bigym) / Delta Joint (for RLBench) \\
 & Temporal ensemble weight & 0.01 (as in ACT \cite{zhao2023learning}) \\
 & Exploration noise ($\epsilon$) & $\text{clip}(\mathcal{N}(0, \sigma), -c, c)$ with $\sigma=0.01$ and $c=0.1$ \\
\bottomrule
\end{tabular}
\caption{The specific structure and hyperparameters of AC3.}
\label{tab:hyperparameters}
\end{table*}

In Figure \ref{fig:model_architecture} and Table \ref{tab:hyperparameters}, we present in detail the specific structure and various hyperparameters of AC3.

\section{Appendix B. Additional Experiment Results}

In this section, we first introduce the 25 tasks involved in the experiments. Successful demonstration videos for each task have been included in the {\it task\_video} file. Then, we conduct a detailed analysis of our two key designs: (1) self-supervised reward shaping and (2) the intra-chunk $n$-step return.

\subsection{Bigym Tasks } 
BiGym \cite{chernyadev2024bigym} focuses primarily on bi-manual mobile manipulation scenarios, a highly challenging setting in robotics. The data is collected via tele-operation, and the demonstrations often exhibit noise and multimodality. These characteristics pose significant challenges for RL algorithms, which must leverage such demonstrations to solve sparsely rewarded tasks.

In our work, we selected 15 tasks from BiGym. All tasks follow a sparse reward scheme: a reward of 1 is granted only upon successful completion, and 0 otherwise.
For observations, we use 84×84 resolution RGB inputs from cameras mounted on the {\it head}, {\it left wrist}, and {\it right wrist}, supplemented with low-dimensional proprioceptive states. For action control, we employ two modes: (i) absolute joint position control; and (ii) a floating base configuration, where locomotion is handled by classic controllers. We use Unitree H1 with two parallel grippers.

Task details are as follows, directly adapted from the BiGym paper \cite{chernyadev2024bigym}:
\begin{enumerate}
    \item {\it Reach Target Dual}: 
    \begin{itemize}
        \item {\it Maximum Episode Length}: 100
        \item {\it Description}: The distance from the {\it left wrist} and the {\it right wrist} to their corresponding goals are smaller than a tolerance value. The default tolerance value is 0.1.
        \item {\it Success Criterion}: The distance from the {\it left wrist} and the {\it right wrist} to their corresponding goals are smaller than a tolerance value. The default tolerance value is 0.1.
    \end{itemize}
    \item {\it Move Plate}:
    \begin{itemize}
        \item {\it Maximum Episode Length}: 300
        \item {\it Description}: The following conditions must be met: (a) the orientation of the plate is upright, (b) the plate is not colliding with the table, (c) the plate is colliding with the rack and (d) the robot has released the plate from its gripper.
        \item {\it Success Criterion}: The following conditions must be met: (a) the orientation of the plate is upright, (b) the plate is not colliding with the table, (c) the plate is colliding with the rack and (d) the robot has released the plate from its gripper.
    \end{itemize}
    \item {\it Move Two Plates}:
    \begin{itemize}
        \item {\it Maximum Episode Length}: 550
        \item {\it Description}: Transfer two plates to the target rack and meet all conditions similar to the move plate task.
        \item {\it Success Criterion}: Transfer two plates to the target rack and meet all conditions similar to the move plate task.
    \end{itemize}
    \item {\it Sandwich Remove}:
    \begin{itemize}
        \item {\it Maximum Episode Length}: 540
        \item {\it Description}: All the following conditions must be met: (a) The sandwich is in collision with the board. (b) The orientation of the sandwich is either up or down.
        \item {\it Success Criterion}: All the following conditions must be met: (a) The sandwich is in collision with the board. (b) The orientation of the sandwich is either up or down.
    \end{itemize}
    \item {\it Sandwich Toast}:
    \begin{itemize}
        \item {\it Maximum Episode Length}: 660
        \item {\it Description}: All the following conditions must be met: (a) The sandwich is in collision with the pan. (b) The orientation of the sandwich is either up or down. (c) The pan is in collision with the hob.
        \item {\it Success Criterion}: All the following conditions must be met: (a) The sandwich is in collision with the pan. (b) The orientation of the sandwich is either up or down. (c) The pan is in collision with the hob.
    \end{itemize}
    \item {\it Sandwich Flip}:
    \begin{itemize}
        \item {\it Maximum Episode Length}: 620
        \item {\it Description}: Similar to {\it sandwich toast}. In addition the sandwich orientation must be flipped.
        \item {\it Success Criterion}: Similar to {\it sandwich toast}. In addition the sandwich orientation must be flipped.
    \end{itemize}
    \item {\it Store Box}:
    \begin{itemize}
        \item {\it Maximum Episode Length}: 600
        \item {\it Description}: The box is in collision with the shelf and the robot has released the box from its grippers.
        \item {\it Success Criterion}: The box is in collision with the shelf and the robot has released the box from its grippers.
    \end{itemize}
    \item {\it Pick Box}:
    \begin{itemize}
        \item {\it Maximum Episode Length}: 540
        \item {\it Description}: The box is in collision with the counter and the robot has released the box from its grippers.
        \item {\it Success Criterion}: The box is in collision with the counter and the robot has released the box from its grippers.
    \end{itemize}
    \item {\it Saucepan To Hob}:
    \begin{itemize}
        \item {\it Maximum Episode Length}: 440
        \item {\it Description}: The saucepan is in collision with the hob and the robot has released the saucepan from its grippers.
        \item {\it Success Criterion}: The saucepan is in collision with the hob and the robot has released the saucepan from its grippers.
    \end{itemize}
    \item {\it Take Cups}:
    \begin{itemize}
        \item {\it Maximum Episode Length}: 420
        \item {\it Description}: All cups are in collision with the counter on the table and the robot has released the cups from its gripper.
        \item {\it Success Criterion}: All cups are in collision with the counter on the table and the robot has released the cups from its gripper.
    \end{itemize}
    \item {\it Put Cups}:
    \begin{itemize}
        \item {\it Maximum Episode Length}: 425
        \item {\it Description}: All cups are in collision with the cupboard shelf and the robot has released the cups from its gripper.
        \item {\it Success Criterion}: All cups are in collision with the cupboard shelf and the robot has released the cups from its gripper.
    \end{itemize}
    \item {\it Wall Cupboard Open}: 
    \begin{itemize}
        \item {\it Maximum Episode Length}: 300
        \item {\it Description}: The joint angle of two doors of the wall cupboard is close to 1 with a tolerance value. The default value is 0.1.
        \item {\it Success Criterion}: The joint angle of two doors of the wall cupboard is close to 1 with a tolerance value. The default value is 0.1.
    \end{itemize}
    \item {\it Wall Cupboard Close}:
    \begin{itemize}
        \item {\it Maximum Episode Length}: 300
        \item {\it Description}: The joint angle of two doors of the wall cupboard is close to 0 with a tolerance value. The default value is 0.1.
        \item {\it Success Criterion}: The joint angle of two doors of the wall cupboard is close to 0 with a tolerance value. The default value is 0.1.
    \end{itemize}
    \item {\it Drawers Open All}:
    \begin{itemize}
        \item {\it Maximum Episode Length}: 480
        \item {\it Description}: The joint angles of all drawers are close to 1 with a tolerance value. The default value is 0.1.
        \item {\it Success Criterion}: The joint angles of all drawers are close to 1 with a tolerance value. The default value is 0.1.
    \end{itemize}
    \item {\it Drawers Close All}:
    \begin{itemize}
        \item {\it Maximum Episode Length}: 200
        \item {\it Description}: The joint angles of all drawers are close to 0 with a tolerance value. The default value is 0.1.
        \item {\it Success Criterion}: The joint angles of all drawers are close to 0 with a tolerance value. The default value is 0.1.
    \end{itemize}
\end{enumerate}

\begin{figure*}[tbp]
\begin{center}
\includegraphics[width=0.95\linewidth]{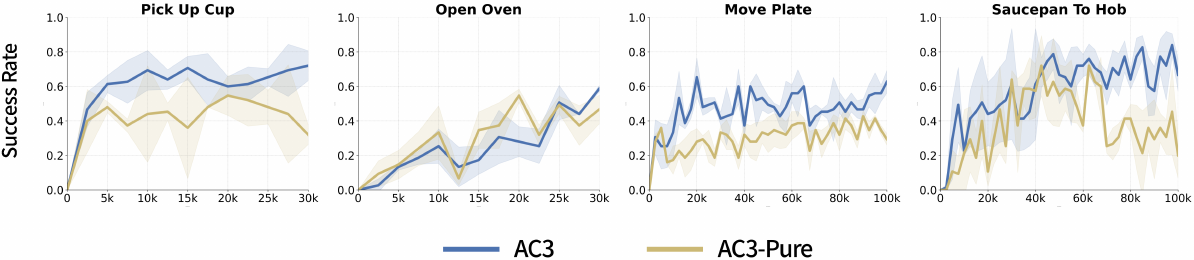}
\end{center}
\caption{Reward shaping effects with extensive experiments. AC3-Pure denotes no reward shaping.}
\label{fig:no_reward}
\end{figure*}

\subsection{RLBench Tasks } 
RLBench \cite{james2020rlbench} is a benchmark focusing on tabletop manipulation. We use RLBench’s dataset generator for collecting 100 synthetic demonstrations for each task; therefore, its trajectories are cleaner compared to Bigym. We select 10 tasks, with the same reward setting as BiGym. We use RGB observations with 84×84 resolution from {\it front}, {\it wrist}, {\it left shoulder}, and {\it right shoulder} cameras. For action control, we employ delta joint position control. We use a 7-DoF Franka Panda robot arm and a parallel gripper.

Task details are as follows:
\begin{enumerate}
    \item {\it Meat On Grill}: 
    \begin{itemize}
        \item {\it Maximum Episode Length}: 150
        \item {\it Description}: Place a piece of meat, either chicken or steak, onto a grill.
        \item {\it Success Criterion}: The meat (chicken or steak) must be detected by the success proximity sensor.
    \end{itemize}
    \item {\it Turn Tap}: 
    \begin{itemize}
        \item {\it Maximum Episode Length}: 125
        \item {\it Description}: Rotate a tap (either left or right).
        \item {\it Success Criterion}: The tap's joint angle reaches 1.57 radians.
    \end{itemize}
    \item {\it Put Money In Safe}: 
    \begin{itemize}
        \item {\it Maximum Episode Length}: 150
        \item {\it Description}: Pick up a stack of money and place it onto a shelf within a safe.
        \item {\it Success Criterion}: The money is no longer grasped by the robot's gripper and is detected by the proximity sensor associated with the target shelf.
    \end{itemize} 
    \item {\it Toilet Seat Up}: 
    \begin{itemize}
        \item {\it Maximum Episode Length}: 150
        \item {\it Description}: Raise the toilet seat.
        \item {\it Success Criterion}: The {\it toilet seat up revolute joint} reaches an angle of 1.40 radians.
    \end{itemize}
    \item {\it Take Plate Off Colored Dish Rack}:
    \begin{itemize}
        \item {\it Maximum Episode Length}: 150
        \item {\it Description}: Pick up a plate from a specific colored dish rack and place it on a surface.
        \item {\it Success Criterion}: The plate is no longer on its initial position on the dish rack, is detected by a {\it success target} proximity sensor, and the robot's gripper is not grasping anything.
    \end{itemize}
    \item {\it Put Rubbish In Bin}:
    \begin{itemize}
        \item {\it Maximum Episode Length}: 150
        \item {\it Description}: Pick up the rubbish and place it inside the bin.
        \item {\it Success Criterion}: The rubbish is detected by the success proximity sensor inside the bin.
    \end{itemize}
    \item {\it Pick Up Cup}:
    \begin{itemize}
        \item {\it Maximum Episode Length}: 100
        \item {\it Description}: Identify and pick up a specifically colored cup from a set of cups.
        \item {\it Success Criterion}: The cup is no longer detected by the success proximity sensor, and the robot's gripper is grasping the cup.
    \end{itemize}
    \item {\it Phone On Base}: 
    \begin{itemize}
        \item {\it Maximum Episode Length}: 175
        \item {\it Description}: Pick up a phone and place it onto its base.
        \item {\it Success Criterion}: The phone is detected by the success proximity sensor, and the robot's gripper is not grasping anything.
    \end{itemize} 
    \item {\it Open Oven}: 
    \begin{itemize}
        \item {\it Maximum Episode Length}: 225
        \item {\it Description}: Open the oven door.
        \item {\it Success Criterion}: The {\it oven door} is detected by the success proximity sensor, and the robot's gripper is not grasping anything.
    \end{itemize}
    \item {\it Put Books On Bookshelf}: 
    \begin{itemize}
        \item {\it Maximum Episode Length}: 175
        \item {\it Description}: Pick up a specified number of books and place them onto the bookshelf.
        \item {\it Success Criterion}: The specified number of books are all detected by the success proximity sensor, and the robot stops at a waypoint.
    \end{itemize}
\end{enumerate}

\subsection{Detailed Analysis of Self-Supervised Reward Shaping }

\noindent{\bf Implementation Details. } The self-supervised reward shaping module is pre-trained using only offline data. Its specific structure, {\it i.e.}, goal network $G_{\omega}(\cdot)$, is identical to that of the state encoder, as shown in Figure \ref{fig:state_encoder}(a).
Specifically, assume a demonstration trajectory $\tau_{0}$ of length $L$. We use all transitions in this trajectory as sequential query samples. Taking timestep $t$ as an example, we extract the state $s_t$ from the current timestep as the query sample $s_q$, take states within a 5-timestep window as positive samples ({\it i.e.}, $s_p \in \{s_{t-5}, ..., s_{t-1}, s_{t+1}, ..., s_{t+5}\}$), and take states in the current trajectory that are $0.1 \times L$ time steps away from it and all states from other trajectories as negative samples ({\it i.e.}, $s_n \in \{s_{0}, ..., s_{t-0.1*L}, s_{t+0.1*L}, ..., s_{L-1}\}$ or $s_{n} \in \tau_{i}, i \neq 0$). For each positive sample, we randomly select 5 negative samples and combine them into positive-negative sample training pairs $[s_q, s_p, s_n]$. Therefore, for each query sample, $10 \times 5 = 50$ training sample pairs can be obtained.

The positive-negative sample selection strategy is repeated randomly 5 times. For each selection, the chosen samples are pre-trained for 5 epochs, thereby preventing the sample encoding of the goal network from overfitting to a single selection of samples. Therefore, the pre-training process consists of a total of 25 epochs. We use triplet loss to train these positive-negative sample pairs:
\begin{equation}
\begin{aligned}
\mathcal{L}_{\omega} =& \mathbb{E}_{(s_q, s_p, s_n) \sim \mathcal{D}_{\text{demo}}} [ \max ( \| G_\omega(s_q) - G_\omega(s_p) \|^2 \\
&- \| G_\omega(s_q) - G_\omega(s_n) \|^2 + m, 0 ) ],
\end{aligned}
\end{equation}
where $s_q$ is the query sample, $s_p$ is the positive sample, $s_n$ is the negative sample, and $m$ is a margin hyperparameter that enforces the distance to the negative sample to be greater than the distance to the positive sample by at least $m$.

\noindent{\bf Efficieny Analysis. }Due to the limited demonstrations, the pre-training process of the goal network is very fast. It can be completed in less than one hour for most tasks, which is far shorter than the training time of the policy network.
Before online RL, we encode all states from the offline demonstrations using the pre-trained goal network and then store them in an array. Similarly, with a small amount of demonstrations, we need less than 10MB of space to store this data.

\begin{figure*}[tbp]
\begin{center}
\includegraphics[width=0.95\linewidth]{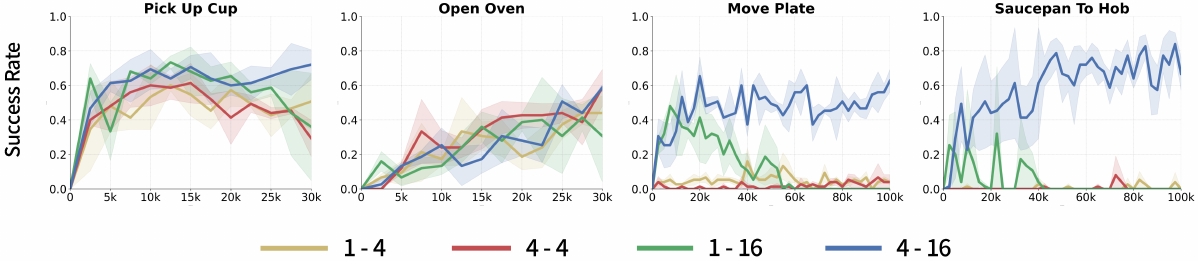}
\end{center}
\caption{A comparison of success rates for various $n$-$C$ settings, where $n$ represents the $n$-step return and $C$ is the action chunk size. More complex bi-manual mobile manipulation tasks are more sensitive to the $n$-$C$ setting, while simple tabletop tasks are relatively insensitive to the choice of $n$-$C$.}
\label{fig:n-C}
\end{figure*}

\begin{figure*}[tbp]
\begin{center}
\includegraphics[width=0.95\linewidth]{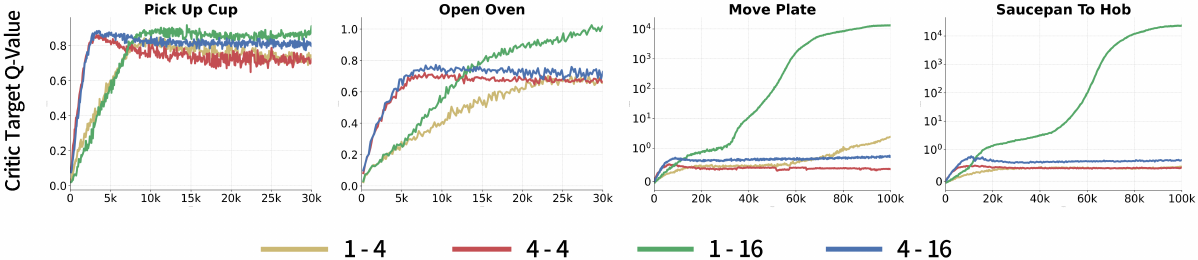}
\end{center}
\caption{The critic target $Q$-values corresponding to the experiments shown in Figure \ref{fig:n-C} (visualize the results from one-run).}
\label{fig:target_q_value}
\end{figure*}

Additionally, we present more extensive experiments on the reward shaping method, including a greater number of tasks, in Figure \ref{fig:no_reward}. We directly name the AC3 without reward shaping as AC3-Pure. From these more extensive experimental results, we find that the use of reward anchors not only improves the performance of the policy model but also enhances the stability of policy training.

\subsection{Detailed Analysis of Intra-Chunk $n$-step Return }

The core of intra-chunk $n$-step return lies in two fundamental issues: \textbf{\textit{Bias-Variance Trade-off}} and \textbf{\textit{Credit Assignment}}. In TD learning, the $n$-step return is a common technique for balancing the bias-variance trade-off between one-step TD ({\it i.e.}, $n=1$) and complete Monte Carlo sampling ({\it i.e.}, $n=\text{len}(\tau)$). Specifically, a smaller value of $n$ leads to a larger bias, while a larger value of $n$ results in a larger variance.

\noindent{\bf Theoretical Analysis of $Q$-value Explosion Problem. } 
When $n<C$, actor-critic-based architecture introduce a new issue, which we refer to as the \textbf{\textit{Unconstrained Action Subspace}} problem. In the case of $n<C$, the action chunk $\mathcal{A}_{t}$ can be decomposed into two parts:
\begin{enumerate}[leftmargin=2em]
    \item The execution head $\mathcal{A}_{t}^{\text{head}}=\{a_t, ..., a_{t + n - 1}\}$: this is the part that directly affects the $n$-step TD target.
    \item The unexecuted tail $\mathcal{A}_{t}^{\text{tail}}=\{a_{t+n}, ..., a_{t+C-1}\}$: this part constitutes the unconstrained action subspace.
\end{enumerate}

When calculating the TD target, the learning target under the $n$-step return setting is 
\begin{equation}
\tilde{y}_t = \sum_{k=0}^{n-1} \gamma^{k}r_{t+k} + \gamma^{n}Q_{\phi'}(s_{t+n}, \pi_{\theta}(s_{t+n})),
\end{equation}
for simplicity, we omit the exploration noise and the double Q-network in the above equation. The key point is that the value of $\tilde{y}_t$ is conditionally independent of the tail of the chunk $\mathcal{A}_{t}^{\text{tail}}$ given the state $s_t$ and the execution head $\mathcal{A}_{t}^{\text{head}}$. This means that regardless of how the tail of the chunk $\mathcal{A}_{t}^{\text{tail}}$ changes, as long as the head $\mathcal{A}_{t}^{\text{head}}$ remains unchanged, both the environmental feedback $(\{r_t, ..., r_{t+n-1}\}, s_{t+n})$ and the TD target will be the same.

\begin{figure*}[tbp]
\begin{center}
\includegraphics[width=0.95\linewidth]{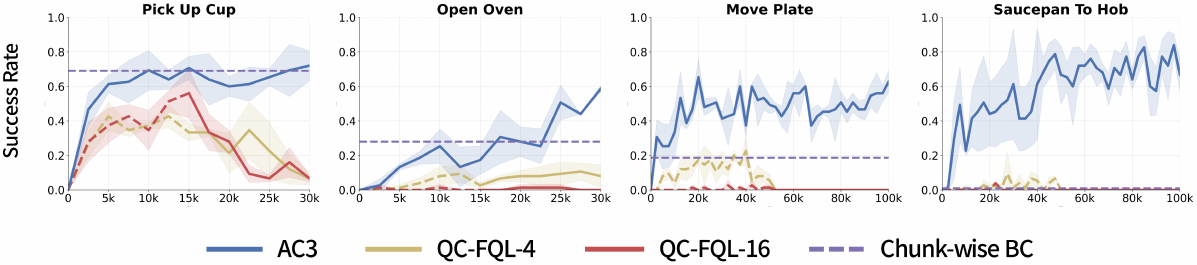}
\end{center}
\caption{QC-FQL results compared with AC3 and chunk-wise BC baseline. QC-FQL-4 indicates a chunk size of 4, and QC-FQL-16 is analogous. }
\label{fig:qc_fql}
\end{figure*}

In this case, the dimension of the unconstrained action subspace is $(C-n)\times d_a$. The actor network can freely manipulate the actions within this subspace to maximize the $Q$-value output without affecting the critic's TD target, and consequently, without direct punishment from the environment. Specifically, the update gradient of the actor can be formulated as:
\begin{equation}
\begin{aligned}
\nabla_{\theta} J(\theta) = \mathbb{E}_{s_t \sim \mathcal{B}} [&\nabla_{\theta}\pi_{\theta}(s_t) \cdot \\
& \nabla_{\mathcal{A}_{t}} Q_{\phi}(s_t, \mathcal{A}_{t})\vert_{\mathcal{A}_{t}=\pi_{\theta}(s_t)}].
\end{aligned}
\end{equation}
The actor's update depends on the critic's gradient with respect to the entire action chunk $\mathcal{A}_t$, denoted as $\nabla_{\mathcal{A}_t}Q_{\phi}$. This gradient can be decomposed into the gradient with respect to the head and that with respect to the tail:
\begin{equation}
\nabla_{\mathcal{A_t}Q_{\phi}} = (\nabla_{\mathcal{A}_{t}^{\text{head}}}Q_{\phi}, \nabla_{\mathcal{A}_{t}^{\text{tail}}}Q_{\phi}).
\end{equation}
Meanwhile, the critic's loss function is $\mathcal{L}_{\phi} = \frac{1}{2} (Q_{\phi}(s_t,\mathcal{A}_t) - \tilde{y}_t)^2$. Its gradient with respect to the parameters $\phi$, according to the chain rule is:
\begin{equation}
\frac{\partial \mathcal{L}(\phi)}{\partial \phi} = \bigl(Q_{\phi}(s_t, \mathcal{A}_t) - \tilde{y}_t\bigr) \cdot \frac{\partial Q_{\phi}(s_t, \mathcal{A}_t)}{\partial \phi}.
\end{equation}
In this gradient formula, $\tilde{y}_t$ acts as the {\it learning target}, guiding the direction of parameter updates. However, as analyzed above, $\mathcal{A}_{t}^{\text{tail}}$ is completely irrelevant to $\tilde{y}_t$. This means there is no information to guide the critic on how to correctly update the $Q$-value with respect to changes in the unconstrained tail $\mathcal{A}_{t}^{\text{tail}}$.

A fatal loop between the actor and critic can thus arise:
\begin{enumerate}[leftmargin=2em]
    \item The critic, training blindly on $\mathcal{A}_{t}^{\text{tail}}$, may generate some steep, meaningless gradients, {\it i.e.}, $\nabla_{\mathcal{A}_{t}^{\text{tail}}}Q_{\phi}$ becomes excessively large.
    \item During its updating, the actor observes this gradient but cannot distinguish between a meaningful and a meaningless gradient, treating it instead as a direction to continuously increase the $Q$-value.
    \item This falsely inflated $Q$-value sample is stored in the replay buffer and reused for training the critic, amplifying the TD error. Consequently, the critic's weights undergo more drastic adjustments, making the gradients even steeper.
    \item The actor thereby identifies an ostensibly better direction to boost the $Q$-value, and this cycle repeats until the $Q$-value explodes.
\end{enumerate}

\noindent{\bf Balancing the Choice between $n$ and $C$. } As analyzed above, under the setting of intra-chunk $n$-step return, a good choice requires an effective trade-off among \textbf{\textit{Bias-Variance Trade-off}}, \textbf{\textit{Credit Assignment}}, and \textbf{\textit{Unconstrained Action Subspace}}. 
\begin{itemize}
    \item In long-horizon, reward-sparse tasks, longer action consistency is a key factor in ensuring the effectiveness of the policy, so the setting of $C$ should not be too small. As shown in Figure \ref{fig:n-C}, the setting of $C = 4$ can only accomplish simple single-arm tabletop tasks, {\it i.e.}, {\it pick up cup} and {\it open oven}, but completely fails in more difficult bi-manual mobile manipulation tasks, {\it i.e.}, {\it move plate} and {\it saucepan to hob}.
    \item Since $C$ should not be too small, a setting of $n = 1$ will lead to an excessively large unconstrained action subspace, thereby causing the problem of $Q$-value explosion and policy failure. As analyzed above, the unconstrained action subspace is proportional to action dimension $d_a$, {\it i.e.}, DoF, which means the $Q$-value explosion problem is more severe in bi-manual manipulation environments. As shown in Figure \ref{fig:target_q_value}, we clearly demonstrate the changes in $Q$-values during training: the $Q$-values grow steadily from the start of training until they explode, at which point the policy completely fails.
    \item While $n = C$ can theoretically eliminate the problem of the unconstrained action subspace, a larger $C$ necessarily entails a larger $n$, which will result in extremely high variance in the network. In environments with a small number of samples and sparse rewards, the model struggles to converge to an effective policy.
    \item To this end, a balanced approach is to use a relatively large $C$ with a moderate $n$, such as $n = 4, 8$ and $C = 16$. Due to the smoothness property of neural networks, a moderate $n$ can also impose a smoothness constraint on the actions in the subsequent $C-n$ timesteps, thereby keeping these actions within a reasonable range. In the main paper, we adopted the setting of $n = 4$ and $C = 16$ across 25 tasks, where we achieved favorable results. This demonstrates that a moderate $n$ is sufficient to suppress the $Q$-value explosion problem. The results presented in Figure \ref{fig:target_q_value} also indicate that a moderate $n$ is sufficient to keep the $Q$-values stable in the later stages of training.
\end{itemize}

\noindent{\bf Special Cases: QC-FQL. } Q-Chunking (QC) \cite{li2025reinforcement} is a concurrent work that proposes a fully chunk-based RL algorithm, {\it i.e.}, where $n=C$. It employs flow matching \cite{lipman2022flow,park2025flow} for behavior cloning pre-training on large-scale offline datasets, followed by online RL using all past experiences. We reproduced the actor-critic version of QC, {\it i.e.}, QC-FQL, using the PyTorch framework, which can also directly output continuous action chunks. For QC-FQL, we construct the {\it actor\_bc\_flow} and {\it actor\_onestep\_flow} networks using a network architecture similar to the actor in AC3, and construct its {\it value} network with an architecture identical to the critic in AC3. We allocated half of the training steps to offline pre-training and the other half to online RL, and the experimental results are presented in Figure \ref{fig:qc_fql}. QC-FQL-4 indicates a chunk size is 4, and QC-FQL-16 is analogous. 

Since the behavior cloning of QC-FQL relies on the flow matching process, it often requires a large amount of offline data to learn a good data distribution. Therefore, with a small number of offline samples, the results of offline training are significantly lower than our chunk-wise BC baseline. In the online RL phase, the use of all experiences for policy updates almost caused policy degradation in all tasks and led to complete failure of the policy in bi-manual tasks ({\it i.e.}, {\it move plate} and {\it saucepan to hob}), which have a larger action space. This shows that with a small amount of offline data and completely sparse rewards, complex behavior cloning methods often fail to achieve the expected results, and the interference of failure experiences on the policy network can be catastrophic.

\end{document}